\pdfoutput=1

\documentclass[11pt]{article}

\usepackage[preprint]{acl}

\usepackage{times}
\usepackage{latexsym}

\usepackage[T1]{fontenc}

\usepackage[utf8]{inputenc}

\usepackage{microtype}

\usepackage{inconsolata}

\usepackage{graphicx}

\usepackage{microtype}
\usepackage{hyperref}
\usepackage{url}
\usepackage{booktabs}
\usepackage{arydshln}
\usepackage{booktabs}

\usepackage{makecell}    
\usepackage{graphicx}    
\usepackage{subcaption}  
\usepackage{algorithm}   
\usepackage{amsthm}
\usepackage{algpseudocode}
\usepackage{xcolor}      
\usepackage{tikz}        
\usepackage{mathtools}   
\usepackage{wrapfig}     
\usepackage{lipsum}      
\usepackage{sidecap}     
\usepackage{pifont}      
\usepackage[flushleft]{threeparttable} 
\usepackage{enumitem}    
\usepackage{diagbox}     
\usepackage{xspace}
\usepackage[T1]{fontenc}
\usepackage[utf8]{inputenc}
\usepackage{microtype}
\usepackage{inconsolata}
\usepackage{colortbl} 
\usepackage{amssymb}
\usepackage{tcolorbox}
\usepackage{multicol}
\usepackage{multirow}

\definecolor{darkblue}{rgb}{0, 0, 0.5}
\hypersetup{colorlinks=true, citecolor=darkblue, linkcolor=darkblue, urlcolor=darkblue}

\title{\model: Enhancing Multimodal Models with\\Mixtures of Multimodal Interaction Experts}

\author{Haofei Yu$^1$\thanks{Equal Contribution.}, Zhengyang Qi$^1$\footnotemark[1], Lawrence Jang$^1$\footnotemark[1],\\
\textbf{Ruslan Salakhutdinov$^1$, Louis-Philippe Morency$^1$, Paul Pu Liang$^2$}\\
$^1$Carnegie Mellon University, $^2$Massachusetts Institute of Technology\\
\texttt{\{haofeiy,zqi2,ljang\}@cs.cmu.edu} \quad \texttt{ppliang@mit.edu} \\
}

\newtheorem{definition}{Definition}

\newtheorem{theorem}{Theorem}

\newcommand{\sarcasm}{\texttt{MUStARD}\xspace}
\newcommand{\model}{{\sc{MMoE}}\xspace}
\newcommand{\mmsarcasm}{\texttt{MMSD}\xspace}
\newcommand{\mmsd}{\texttt{MMSD2.0}\xspace}
\newcommand{\urfunny}{\texttt{URFunny}\xspace}

\begin{document}

\maketitle

\begin{abstract}
Advances in multimodal models have greatly improved how interactions relevant to various tasks are modeled. Today's multimodal models mainly focus on the correspondence between images and text, using this for tasks like image-text matching. However, this covers only a subset of real-world interactions. Novel interactions, such as sarcasm expressed through opposing spoken words and gestures or humor expressed through utterances and tone of voice, remain challenging. In this paper, we introduce an approach to enhance multimodal models, which we call \textbf{M}ultimodal \textbf{M}ixtures \textbf{o}f \textbf{E}xperts (\model). The key idea in \model\ is to train separate expert models for each type of multimodal interaction, such as redundancy present in both modalities, uniqueness in one modality, or synergy that emerges when both modalities are fused. On a sarcasm detection task (\sarcasm) and a humor detection task (\urfunny), we obtain new state-of-the-art results. \model\ is also able to be applied to various types of models to gain improvement.
\end{abstract}

\vspace{-3mm}
\section{Introduction}
\vspace{-1mm}

\begin{figure}[t!]
    \centering
    \includegraphics[scale=0.59]{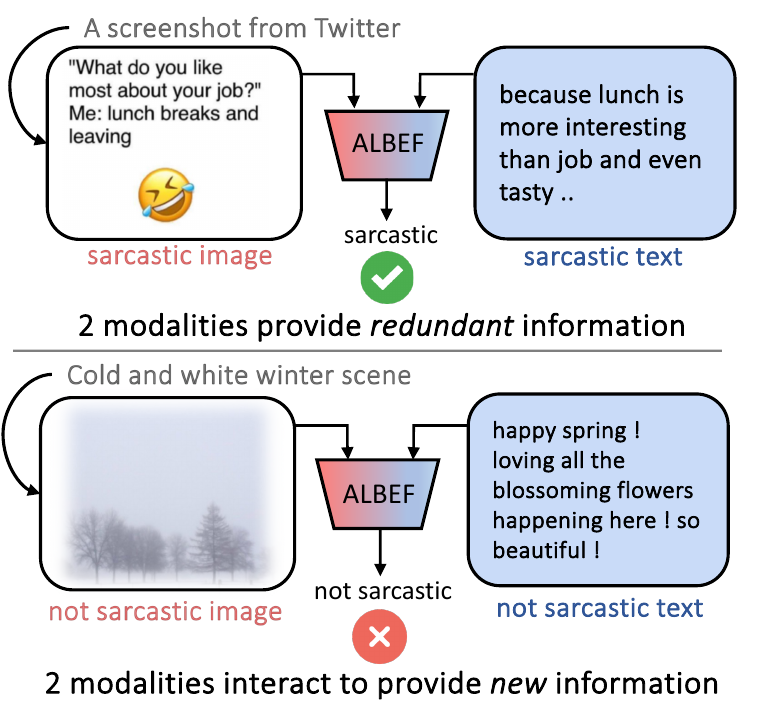}
    \vspace{-2mm}
    \caption{\textbf{A single model cannot handle all types of multimodal interactions well for hard multimodal prediction tasks.} For example, to predict sarcasm, ALBEF can have $\sim$89\% F1 when modalities contain redundant information (e.g., both the text and the image are sarcastic), but drops to $\sim$24\% F1 when there is synergy between modalities (e.g., the image shows a cold winter scene and the text says it is a happy spring, indicating the user's sarcastic intent about the weather).}
    \label{fig:intro}
    \vspace{-4mm}
\end{figure}

Recent advances in the design and pretraining of vision-language models have enabled significant progress in capturing the correspondences between images and text~\citep{zhu2023minigpt,li2023blip,liu2023visual}. These models have seen successes in image captioning~\citep{xu2015show}, text-to-image generation~\citep{saharia2022photorealistic}, multimodal retrieval~\citep{mithun2018learning}, multimodal classification~\citep{li2021align}, and more. At its core, these methods aim to capture overlaps in semantic content between images and text, making a strong multi-view redundancy assumption~\citep{tian2020makes,liang2023factorized,pmlr-v139-zbontar21a}. However, redundancy is only one type of interaction seen between two modalities~\citep{williams2010nonnegative,liang2023quantifying,marsh2003taxonomy}. Instead, it might hinge on \textit{unique} details from either modality (e.g. detecting laughter from someone not observed) or the result of \textit{synergistic} fusion of both modalities, producing insights absent when either modality is considered in isolation (e.g., sarcasm and humor discerned from incongruent speech and gestures). Synergy is particularly interesting because it often arises when the predictions from different modalities are \textit{contradicting}, or \textit{incongruent} with one another~\citep{bateman2014text,kruk2019integrating,zhang2018equal}.

The diversity of possible real-world multimodal interactions poses a challenge to today's multimodal models. Empirically, we find that \textit{one single model may not be the most suitable in capturing all types of interaction at the same time}. For example, models trained to learn the correspondences between words and image regions (e.g., for retrieval) will struggle when there is unique information in one modality~\citep{liang2023factorized,winterbottom2020modality}, or when the image and text provide contradicting information that must be contextualized together~\citep{hessel2022androids}. We show an example of this failure in Figure~\ref{fig:intro}, where ALBEF~\citep{li2021align} can easily detect sarcasm when it is present in both modalities (redundancy), but fails when the sarcastic intent arises from the synergistic fusion of both image and text. Quantitatively, ALBEF has a performance drop of up to 60\% on data with synergistic interactions compared with those with redundancy interactions.

To tackle this problem, we propose \model, by leveraging the key insight that different interactions require different modeling paradigms. A natural way to model these differences is to use a mixture of multimodal experts with one specialized expert model for each interaction.
Each expert model can be specialized based on the unique training data they see or a special training objective. Furthermore, there is evidence that the brain also uses separate expert regions during the multisensory integration process, depending on the types of input modalities and multimodal contexts present during perception~\citep{stein2020multisensory}.
During inference on unseen data points at testing time, \model\ relies on specific fusion methods to provide weights for each expert model, combine the output of each expert model, and obtain a final prediction.

\model\ achieves new state-of-the-art results on one multimodal sarcasm detection dataset and one multimodal humor detection dataset we tested on, \sarcasm and \urfunny. Moreover, we show that our approach is easy to implement on various types of models: fusion-based vision language models like ALBEF~\citep{li2021align}, multimodal extended large language models like BLIP2~\citep{li2023blip}, and image-captioned large language models like Qwen2~\citep{qwen2} all improve after adding \model on top of them. \footnote{Codebase and reproduction guidance are available at \href{https://github.com/lwaekfjlk/mmoe}{https://github.com/lwaekfjlk/mmoe}}

\vspace{-1mm}
\section{Related Work}
\vspace{-1mm}

We cover related work in quantifying and learning multimodal interactions, as well as recent advances in multimodal large language models, ensembling, and mixtures of experts.

\paragraph{Multimodal Interactions} defines the degrees of commonality between modalities and the ways they combine to provide new information for a task~\citep{liang2022foundations}. A core problem lies in understanding the nature of how modalities interact and modeling these interactions using data-driven methods. The study of multimodal interactions has involved semantic definitions based on research in multimedia~\citep{marsh2003taxonomy}, human (and animal) communication~\citep{partan2005issues,flom2007development,ruiz2006examining}, and human social interactions~\citep{mai2019divide, jung2018multi}. These have also inspired statistical methods to quantify multimodal interactions from unimodal predictions~\citep{mazzetto21a}, trained model weights and activations~\citep{sorokina2008detecting,tsang2018detecting,tsang2019feature,hessel2020emap}, feature selection~\citep{ittner2021feature,yu2003efficiently,yu2004efficient,auffarth2010comparison}, and information theory~\citep{liang2023quantifying,liang2023multimodal,williams2010nonnegative,bertschinger2014quantifying}. Our work builds on this line of work in quantifying multimodal interactions, particularly the statistical definitions that enable accurate estimation from large-scale multimodal datasets.

\begin{figure*}[t!]
    \centering
    \includegraphics[scale=0.55]{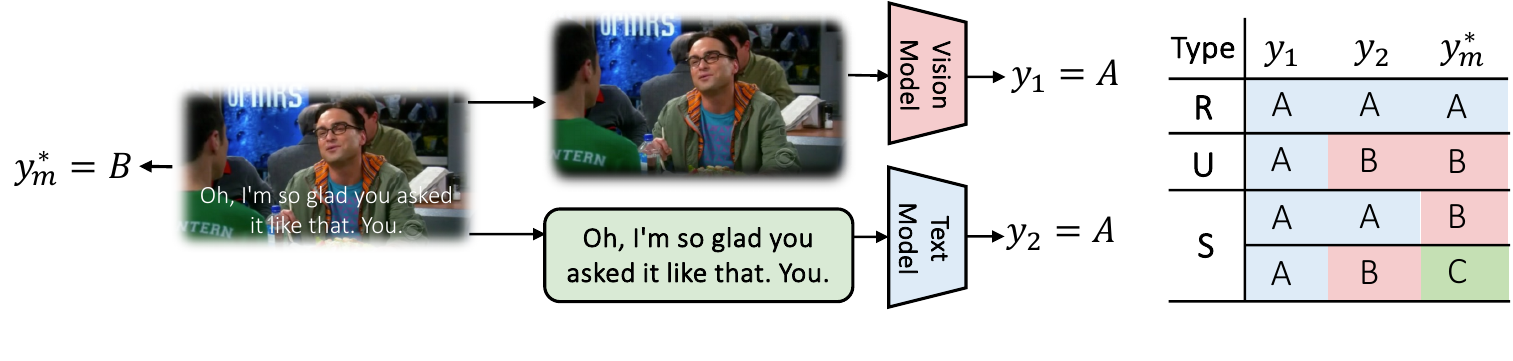}
    \vspace{-0mm}
    \caption{\textbf{We classify one multimodal dataset into three subsets based on their multimodal interactions}: (1) \textit{Redundancy} (R), when both modalities provide the same prediction, (2) \textit{Uniqueness} (U), when two modalities make different predictions, of which one of them is correct, (3) \textit{Synergy} (S), when the ground-truth multimodal labels do not agree with either of unimodal predictions.
    $y_1$ represents the prediction based on vision modality, $y_2$ represents the prediction from text modality, and $y_m^{*}$ represents the ground-truth label. \{$A$, $B$, $C$\} represents classes.}
    \label{fig:classification}
    \vspace{-0mm}
\end{figure*}

\vspace{-1mm}
\paragraph{Multimodal Language Models} have revolutionized multimodal learning since representations of images and text can now be fed into large language models for flexible question-answering, reasoning, and multi-turn dialog conditioned on images. Many of these models are built on top of multimodal extensions of the Transformer architecture~\citep{su2019vl,liang2022highmmt,jaegle2021perceiver,lu2019vilbert,tsai2019multimodal,tan2019lxmert}. In addition to training large-scale multimodal transformers \textit{natively} from input modalities, another line of work takes pre-trained language and vision models and aims to learn a small set of \textit{adapter} parameters to align visual and language representations~\citep{koh2023grounding,li2023blip,zhu2023minigpt}. These approaches have shown strong performance on many multimodal benchmarks, such as in visual question answering~\citep{wang2022ofa}, text-to-video generation~\citep{kondratyuk2023videopoet}, robotics tasks~\citep{driess2023palm}, and biomedical analysis~\citep{acosta2022multimodal}. However, these methods train monolithic models that perform the same computation for all types of multimodal interactions, which we show to be suboptimal and inefficient when datasets contain a mix of diverse and complex interactions.

\paragraph{Ensembles and Mixtures of Experts} are commonly used techniques to boost a model's performance using a collection of expert models each with their specialized expertise but individually weaker than baseline~\citep{freund1996experiments}. \citet{cheng2020voting} utilized a voting-based method to ensemble predictions from multiple models to provide more accurate answers. Besides discrete voting, continuous ensembles in logit space have also been proposed~\citep{eigen2013learning,tasci2021voting}. In settings where it is difficult to define which expert is correct, trainable ensemble functions have been designed to automatically combine multiple experts in an end-to-end fashion~\citep{he2021fastmoe,shazeer2017outrageously,du2022glam}. Our work uses these ideas as a foundation to learn different types of multimodal interactions.

\vspace{-1mm}
\section{\textbf{M}ultimodal \textbf{M}ixtures \textbf{o}f \textbf{E}xperts}
\vspace{-1mm}

We focus on multimodal prediction tasks: given feature vectors from two modalities with $x_1$ and $x_2$, our goal is to predict the label $y$ using both $x_1$ and $x_2$. Naturally, task-related information may be contained uniquely in one of the modalities, present redundantly in both, or require synergistically combining of information from both modalities. While prior work has focused on designing a single multimodal model for all data points in a task, our key insight is that each data point may exhibit a different type of interaction and therefore require a different modeling approach. Our method, which we call \model, is a natural solution to this problem in three steps (1) \textit{Categorizing}: categorizing multimodal interaction types in each data point for the training set, (2) \textit{Training}: training three expert models to master at each type of interactions (redundancy, uniqueness, and synergy), (3) \textit{Inference}: dynamically ensembling the mixture of expert models during inference on unseen new data points. We now explain each of these three steps in detail.

\subsection{Categorizing Multimodal Interactions}
\label{sec:classify-interaction}

Prior work has provided definitions of \textit{redundant}, \textit{unique}, and \textit{synergistic} interactions using the language of information theory~\citep{williams2010nonnegative,liang2023quantifying}. However, estimating information theoretic measures can be challenging for high-dimensional and continuous distributions~\citep{perez2008estimation}. When these interactions cannot be exactly computed, they can be approximately inferred by considering whether unimodal models trained on each modality \textit{agree} or \textit{disagree} with each other. We can formalize the concept of modality agreement and disagreement with a discrepancy function as follows:

\vspace{1mm}
\begin{definition}[Prediction Discrepancy Function]
Given feature \( x_1 \in \mathcal{X}_1 \) and \( x_2 \in \mathcal{X}_2 \), and unimodal classifiers \( f_1: \mathcal{X}_1 \rightarrow \mathcal{Y} \) and \( f_2: \mathcal{X}_2 \rightarrow \mathcal{Y} \), let \( y_1 = f_1(x_1) \) and \( y_2 = f_2(x_2) \) denote their predictions. We define the prediction discrepancy function \( \delta(y_1, y_2) \) as a mapping \( \delta: \mathcal{Y} \times \mathcal{Y} \rightarrow \mathbb{R}_{\geq 0} \) that quantifies the dissimilarity between the predictions of \( f_1 \) and \( f_2 \). For tasks with a discrete label space \( \mathcal{Y} \), the discrepancy function is defined as:
\[
\delta(y_1, y_2) = 
\begin{cases} 
0, & \text{if } y_1 = y_2, \\ 
1, & \text{if } y_1 \neq y_2.
\end{cases}
\]
\end{definition}

\begin{figure*}[ht]
    \centering
    \begin{minipage}[t]{0.31\textwidth}
        \centering
        \includegraphics[width=0.87\textwidth]{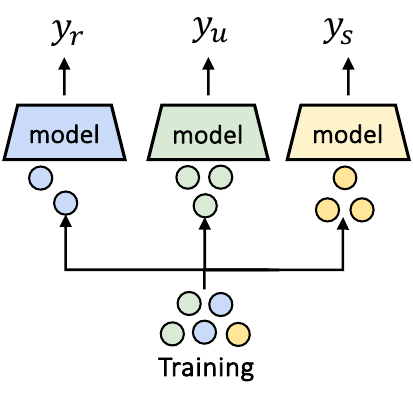}
        \vspace{-3mm}
        \caption{\textbf{\model training}: Each multimodal datapoint is categorized based on its multimodal interaction and used to train an expert model tailored only for that interaction.}
        \vspace{-2mm}
        \label{fig:train-process}
    \end{minipage}%
    \hfill
    \begin{minipage}[t]{0.31\textwidth}
        \centering
        \includegraphics[width=0.87\textwidth]{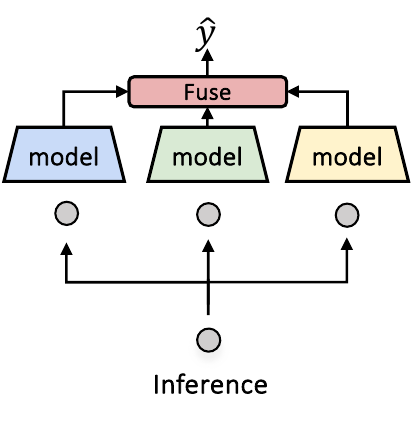}
        \vspace{-3mm}
        \caption{\textbf{\model inference}: We infer which multimodal interaction a test datapoint requires and use a soft weighted fusion over on the outputs from multiple expert models.}
        \vspace{-2mm}
        \label{fig:inference-process}
    \end{minipage}%
    \hfill
    \begin{minipage}[t]{0.31\textwidth}
        \centering
        \includegraphics[width=0.87\textwidth]{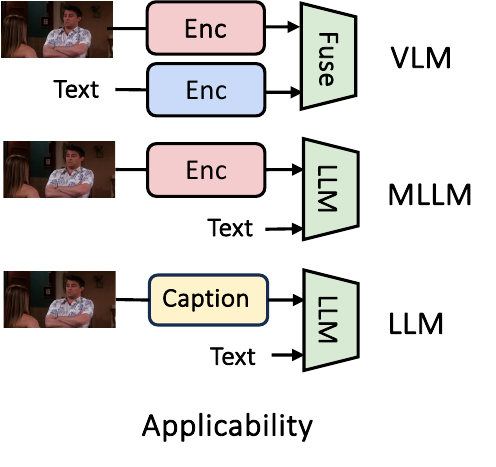}
        \vspace{-3mm}
        \caption{\textbf{\model applicability}: \model\ can be used as a drop-in method to the training of fusion-based VLMs, multimodal extended LLMs, and image-captioned LLMs.}
        \vspace{-2mm}
        \label{fig:model-type}
    \end{minipage}
    \label{fig:combined}
\end{figure*}

The binary discrepancy function indicates that modalities \textit{agree} with each other when $\delta=0$ and modalities \textit{disagree} with each other when $\delta=1$. Combining them with multimodal predictions, gives us an intuitive guideline to categorize data points based on three types of interactions:
\begin{enumerate}[noitemsep,topsep=0pt,nosep,leftmargin=*,parsep=0pt,partopsep=0pt]
    \item \textit{Redundancy}: when both modalities \textit{agree} with the multimodal prediction, two modalities contain redundant information. 
    \item \textit{Uniqueness}: when two modalities \textit{disagree} and one of them is aligned with the multimodal prediction, two modalities contain unique information and one is dominant.
    \item \textit{Synergy}: when the multimodal prediction \textit{disagrees} with both unimodal predictions so there is synergistic information generated from two modalities when predicting.
\end{enumerate}

With such intuitive guidelines above, we formally define the categorization process as follows:

\begin{theorem}[Multimodal Interaction-based Categorization]
Let \( y_1 \) and \( y_2 \) denote the predictions from unimodal classifiers, and let \( y_m \) represent the multimodal prediction from multimodal models. The interaction discrepancy between the predictions is defined as:
\[
\Delta_{1,2}(y_1, y_2, y_m) = \delta(y_1, y_m) + \delta(y_2, y_m)
\]

\noindent where \( \delta(\cdot, \cdot) \) denotes the discrepancy function between two predictions. The categorization is then described as follows:
\setlength{\leftmargini}{12pt}
\begin{itemize}
    \item \( \Delta_{1,2} = 0 \): \textit{Redundancy}, i.e., \( y_1 = y_2 = y_m \),
    \item \( \Delta_{1,2} = 1 \): \textit{Uniqueness}, i.e., \( y_1 = y_m \neq y_2 \) or \( y_2 = y_m \neq y_1 \),
    \item \( \Delta_{1,2} = 2 \): \textit{Synergy}, i.e., \( y_1 \neq y_m \) and \( y_2 \neq y_m \).
\end{itemize}
\end{theorem}

To illustrate the categorization rule, Figure~\ref{fig:classification} shows an example. In practice, obtaining high-quality predictions can be challenging. Labels from multimodal datasets, which are typically generated by humans making multimodal predictions, can be directly used. Also, we obtain high-quality unimodal predictions $y_1$ and $y_2$ via state-of-the-art foundation models in the few-shot prompting style for all training data points. For vision-only predictions, we utilize vision-language models like CogVLM2~\citep{wang2023cogvlm} to obtain them by providing only the query and the image and make sure that generated answers are conditioned only on the vision-side information. To get text-only predictions, we use state-of-the-art language models like Qwen2-72B-Instruct~\citep{qwen2} with the query and the language information so the model answers are conditioned only on text for prediction. More information related to the collection of unimodal labels is available in Appendix \S\ref{appendix:unimodal-label}.

\subsection{Training Expert Models for Each Multimodal Interaction Type}
\label{sec:training-expert-model}

Given the categorization of multimodal datasets into subsets each with a similar interaction, this section describes how we use these interaction-specific subsets to train interaction-specific expert models. Illustrated in Figure \ref{fig:train-process}, there are three specialized models, which we term $f_r, f_u,$ and $f_{s}$ for expert models of redundancy, uniqueness, and synergy respectively. While these individual expert models share the same format of inputs with image and text data pairs, their learning outcomes can differ significantly due to the multimodal data distributions they are trained on.

Overall, for expert model training, we collect all high-quality evidence of redundant interactions to train a redundancy expert model $f_r$. This process is repeated for unique and synergistic interactions, resulting in trained expert models $f_{r}, f_{u}$, and $f_{s}$. Each expert is trained only on the subset of data points that \textit{maximally exhibit that interaction}; this specialization enables experts to be performant at learning that specific interaction. More technical details about the training process of expert models are further discussed in Appendix \S\ref{expert-model-train-detail}.

We also note that it is possible to design interaction expert models using different modeling architectures and training objectives based on innovations in multimodal machine learning. For example, it has been empirically demonstrated that late fusion models are more suitable when modalities are redundant~\citep{gadzicki2020early}, and models with expressive higher-order interactions (e.g., polynomials and tensors) are suitable when there is synergy between modalities~\citep{hou2019deep}. Moreover, multi-task training allows us to leverage the power of scale and learn interaction expert models adaptable to multiple tasks simultaneously. We leave these explorations for our future work.

\subsection{Inference with Mixtures of Expert Models}

The conclusion of Section \S\ref{sec:training-expert-model} yields three expert models each suited for a certain type of multimodal interactions. During inference on unseen test data points, we need to select one or more expert models that are most suitable to get the final prediction. This is a challenge since the categorization of data points during training (presented in Section \S\ref{sec:classify-interaction}) relies on the multimodal prediction $y_m$, which we have during training but not during inference. Therefore, we need to design a method to provide an accurate estimation of the potential multimodal interactions included in one data point.

Our key assumption is that categorizing multimodal interactions within a data point is an essential sub-task that must be completed before the model can generate a final prediction. The multimodal interaction type captures the information shift between unimodal and multimodal inputs. These interaction-type predictions emphasize more general features compared to those needed for task label prediction. Consequently, even if a multimodal model struggles to accurately predict the task label $y^*$, it may still be able to determine the interaction type of the data point (e.g., whether two modalities provide similar, distinct, or synergistic information). This distinction becomes particularly relevant when the prediction task involves regression or classification with many classes. 

Therefore, we \textit{approximately categorize} data during inference through a \textit{soft mixture of weights}, defined as $w_r, w_{u},$ and $w_{s}$ over the three interaction types. These weights are inferred dynamically for each data point using a finetuned fusion model (e.g., BLIP2 in practice). We also test simple model-free baselines like prior constants based on the frequency statistics of each interaction to weight each expert model and so on; see detailed ablation studies on these fusion methods in Section \S\ref{sec:ablation-study} and fusion model training details in Appendix \S\ref{fusion-train}. Using these inferred weights for each expert model, we obtain a final prediction $\hat{y} = \sum_{i=\in \{  r, u, s\}} w_i f_i (x_1,x_2)$ as the output of \model.

\section{Experiments}
\label{sec:experiments}
\vspace{-1mm}

Our experiments are designed to evaluate the effectiveness of our method when applied to a diverse set of multimodal foundation model architectures and multimodal prediction tasks.

\subsection{Experimental Setup}

We introduce the models and multimodal prediction tasks that we consider for experiments in this section. More information related to experimental settings is available in Appendix \S\ref{appendix:exp-details}.

\paragraph{Model} \label{model-type} We implement \model\ on top of three categories of multimodal language models to show its widespread applicability on top of many base models (see Figure \ref{fig:model-type} for an illustration). Detailed model information is available in Appendix \S\ref{asset}. These three model categories include:
\begin{enumerate}[noitemsep,topsep=0pt,nosep,leftmargin=*,parsep=0pt,partopsep=0pt]
    \item \textit{Fusion-based vision language models (VLM)} uses cross-attention to learn multimodal interactions between all regions of the image with all words in the input text. Examples of such models include ALBEF~\citep{li2021align}, LXMERT~\citep{tan2019lxmert} and BLIP~\citep{li2022blip}.
    \vspace{1mm}
    \item \textit{Multimodal-extended LLMs (MLLM)} includes models like BLIP2~\citep{li2023blip} and FROMAGe~\citep{koh2023grounding}. It starts with an image encoder and an LLM as the backbone of the architecture. Most state-of-the-art models are based on multimodal-extended LLMs.
    \vspace{1mm}
    \item \textit{Image-captioned LLMs (LLM)} convert images to text using an image captioning model and uses a text-only LLM like Qwen2~\citep{qwen2} on the concatenation of captioned images and text inputs. Examples include the Socratic Model~\citep{zeng2022socratic} and the video understanding model~\citep{zhang2023simple}.
\end{enumerate}

\paragraph{Multimodal prediction task} We implement our method on three multimodal prediction tasks, including two sarcasm detection tasks, which are \sarcasm~\citep{castro2019towards} and \mmsd~\citep{qin-etal-2023-mmsd2}, and one humor detection task, which is \urfunny~\citep{hasan2019ur}. These tasks require interaction learning to conduct prediction. Detailed information about dataset statistic information and their preprocessing methods are available in Appendix \S\ref{dataset-info} and \S\ref{data-preprocess}.

\vspace{-1mm}
\subsection{Main Results}
\vspace{-1mm}

In this section, firstly, we study how our best \model\ models compare to state-of-the-art baselines on multiple multimodal prediction tasks. Secondly, we study whether \model\ improves performance when applied on top of all three types of base models mentioned in Section \S\ref{model-type}.

\vspace{-1mm}
\paragraph{Overall comparison with state-of-the-art} In Table \ref{tab:main_result}, we show that \model can improve the state-of-the-art performance on both the \sarcasm and \urfunny datasets. Specifically, we outperform LF-DNN-v1~\citep{ding2022multimodal} on the \sarcasm dataset, achieving a 1.35-point improvement in F1 score. On the \urfunny dataset, our fine-tuned BLIP2 model with \model surpasses FDMER~\citep{yang2022disentangled} with a 0.84-point gain in accuracy. 

\begin{table}[t!]
    \centering
    \small 
    \begin{tabular}{@{}p{0.1cm}p{3.8cm}p{1.1cm}l@{}}
        \toprule[1.1pt]
        & \textbf{Model}       & \textbf{Acc} ($\uparrow$) & \textbf{F1} ($\uparrow$) \\
        \midrule
        \multirow{13}{*}{\rotatebox{90}{\textbf{\sarcasm}}} & MulT$\dagger$~\citep{tsai2019multimodal} & - & 64.49  \\
        & LMF$\dagger$~\citep{liu2018efficient} & - & 69.92 \\
        & LFDNNv1$\dagger$~\citep{ding2022multimodal} & - & 70.99  \\
        \cmidrule{2-4} 
        & ALBEF & 54.49$_{\pm 3.13}$ & 48.51$_{\pm 2.21}$ \\ 
        & ALBEF+\model & \underline{54.49}$_{\pm 2.85}$ & \underline{51.95}$_{\pm 2.81}$ \\ 
        \cmidrule{2-4} 
        & BLIP2 & 53.75$_{\pm 9.33}$ & 62.65$_{\pm 2.67}$ \\ 
        & BLIP2+\model & \underline{59.18}$_{\pm 2.11}$ & \underline{64.74}$_{\pm 2.49}$ \\ 
        \cmidrule{2-4} 
        & Qwen2-0.5B & \underline{54.59}$_{\pm 4.35}$ & 58.17$_{\pm 0.86}$ \\ 
        & Qwen2-0.5B+\model & 49.06$_{\pm 3.00}$ & \underline{59.77}$_{\pm 0.35}$\\
        \cmidrule{2-4} 
        & Qwen2-1.5B & 64.79$_{\pm 4.11}$ & 65.38$_{\pm 5.16}$\\ 
        & Qwen2-1.5B+\model & \textbf{\underline{70.69}}$_{\pm 3.28}$ & \underline{\textbf{72.34}}$_{\pm 1.50}$ \\        
        \midrule \midrule
        \multirow{10}{*}{\rotatebox{90}{\textbf{\urfunny}}} & MulT$\dagger$~\citep{tsai2019multimodal} & 66.65 & -  \\
        & FDMER~\citep{yang2022disentangled} & 70.43  & - \\
        \cmidrule{2-4} 
        & ALBEF & 66.77$_{\pm 0.86}$ & 68.67$_{\pm 0.18}$ \\ 
        & ALBEF+MMoE & \underline{67.91}$_{\pm 0.31}$ & \underline{69.85}$_{\pm 0.32}$ \\ 
        \cmidrule{2-4} 
        & BLIP2 & 70.43$_{\pm 0.99}$ & 74.31$_{\pm 0.04}$ \\ 
        & BLIP2+MMoE & \underline{\textbf{71.27}}$_{\pm 0.87}$ & $\underline{\textbf{74.32}}_{\pm 0.05}$ \\ 
        \cmidrule{2-4} 
        & Qwen2-0.5B & \underline{69.29}$_{\pm 0.81}$  & \underline{70.46}$_{\pm 0.14}$ \\ 
        & Qwen2-0.5B+MMoE & 69.19$_{\pm 0.64}$ & 68.38$_{\pm 1.55}$ \\ 
        \midrule \midrule
        \multirow{11}{*}{\rotatebox{90}{\textbf{\mmsd}}} & DynRT-Net~\citep{tian-etal-2023-dynamic} & 71.40 & 71.34 \\
        & MCLIP~\citep{qin-etal-2023-mmsd2} & \textbf{85.64} & 84.10 \\
        & LLaVA-1.5~\citep{liu2024improved} & 85.18 & \textbf{85.11} \\
        \cmidrule{2-4} 
        & ALBEF & 81.79$_{\pm 0.24}$ & 79.33$_{\pm 0.79}$ \\ 
        & ALBEF+\model & \underline{82.30}$_{\pm 0.27}$ & \underline{80.63}$_{\pm 0.52}$ \\
        \cmidrule{2-4} 
        & BLIP2 & 84.75$_{\pm 0.20}$ & \underline{83.52}$_{\pm 0.35}$  \\
        & BLIP2+\model & \underline{84.82}$_{\pm 0.30}$ & 83.38$_{\pm 0.36}$ \\
        \cmidrule{2-4} 
        & Qwen2-0.5B & 81.87$_{\pm 0.54}$ & 80.17$_{\pm 0.14}$\\ 
        & Qwen2-0.5B+\model & \underline{82.27}$_{\pm 0.14}$ & \underline{80.67}$_{\pm 0.20}$ \\
        \bottomrule[1.1pt]
    \end{tabular}
    \caption{\textbf{\model can beat state-of-the-art models for \sarcasm and \urfunny. It can be applied to any type of model for improvement}. The numbers in the table represent the mean values from 3 runs with 3 seeds, with the corresponding standard variance provided. Full results can be found in Appendix \S\ref{additional-exp-result}. $\dagger$ indicates that models utilize all audio, text, and vision information provided in the dataset while ours only utilizes text and vision information for prediction.}
    \label{tab:main_result}
    \vspace{-6mm}
\end{table}

\vspace{-1mm}
\paragraph{Improvement on various types of models}

We first compare the performance of 3 types of models with and without \model on \sarcasm dataset. As shown in Table \ref{tab:main_result}, all models, including ALBEF, BLIP2, and Qwen2, show improvements in F1 scores. Notably, Qwen2-1.5B achieves an increase of 6.96 points, establishing it as the state-of-the-art model on this task. Additionally, on the \mmsd dataset, both ALBEF and Qwen2 demonstrate performance gains, while BLIP2 remains relatively unchanged. For the \urfunny dataset, ALBEF improves accuracy by 1.14 points, and BLIP2 by 0.84 points, whereas Qwen2 experiences a slight decline after applying \model. The performance drop on \urfunny may be due to the inability of image captioning models to provide useful descriptions relevant to humor detection from the TED talk videos. As a result, text-based models like Qwen2 might struggle to achieve further improvements.

Furthermore, when comparing the performance across the three prediction tasks and three models, we observe a general trend: incorporating the \model tends to provide more robust improvements on challenging datasets (e.g., \sarcasm) and weaker models (e.g., ALBEF) with low F1 scores, which initially have lower performance. In contrast, the improvements are less pronounced on easier datasets (e.g., \mmsd) or stronger models (e.g., BLIP2), which already exhibit strong performance.

\vspace{-1mm}
\section{Analysis}
\vspace{-1mm}

Based on these quantitative results, we further provide a fine-grained analysis of our method. First, we examine the limitations of current multimodal models by presenting empirical evidence where a single model faces challenges in typical types of interactions. We then explore whether specialized multimodal interaction expert models excel in their respective interaction types. Furthermore, we analyze the scaling law of expert models and discuss whether these expert models can be potentially smaller, in contrast to typically overparameterized models. Lastly, we provide additional details on the unimodal predictions and emphasize their important role in the data categorization process.

\begin{figure}[t!]
    \centering
    \hspace*{-3mm}
    \includegraphics[scale=0.24]{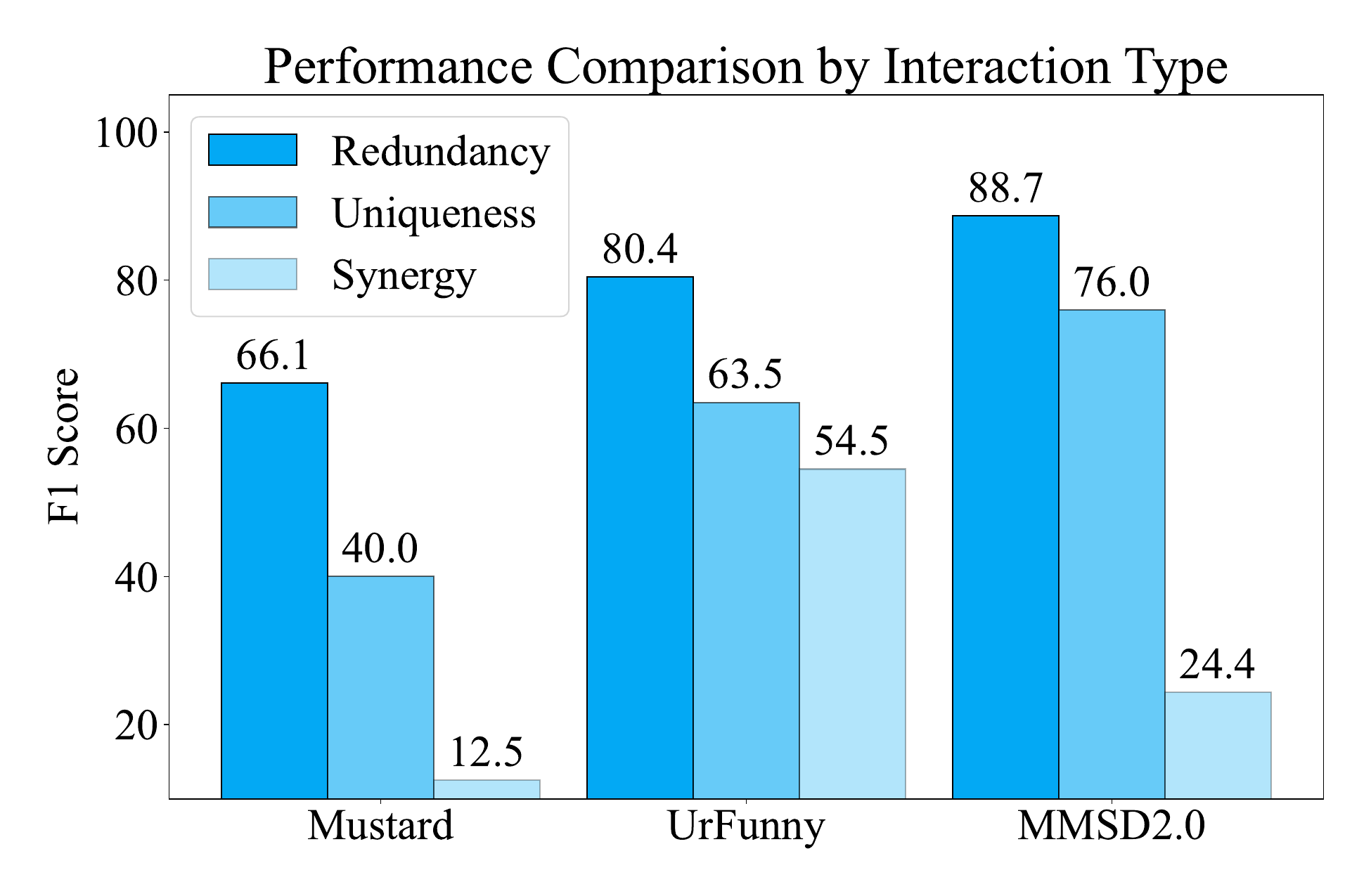}
    \vspace{-8mm}
    \caption{\textbf{Multimodal models struggle with synergy much more than redundancy and uniqueness}. ALBEF shows significantly lower performance on synergistic datapoints compared with redundancy and uniqueness that are categorized based on our method.}
    \label{fig:rus_performance}
    \vspace{-2mm}
\end{figure}  

\subsubsection*{RQ1. What types of multimodal interaction do current models struggle with?}

In Figure \ref{fig:rus_performance}, we categorize all test data points based on their corresponding interaction type using the method mentioned in Section \S\ref{sec:classify-interaction}. We observe significant performance variations when using the same model to predict across data with different interaction types. Across the three datasets—\sarcasm, \urfunny, and \mmsd—data points with synergy interactions show markedly lower F1 scores compared to those with uniqueness interactions, with performance gaps of 27.5, 9.0, and 51.6 for \sarcasm, \urfunny, and \mmsd, respectively. Also, data points with uniqueness interaction perform substantially worse than those with redundancy interaction, with gaps of 26.1, 16.9, and 12.7 for three datasets. These trends are not limited to ALBEF, as we observe similar patterns in BLIP2 and Qwen2, highlighting that data points with strong synergy interactions represent a common challenge across all three types of models.

\begin{figure}[t]
    \centering
    \vspace{3mm}
    \includegraphics[scale=0.4]{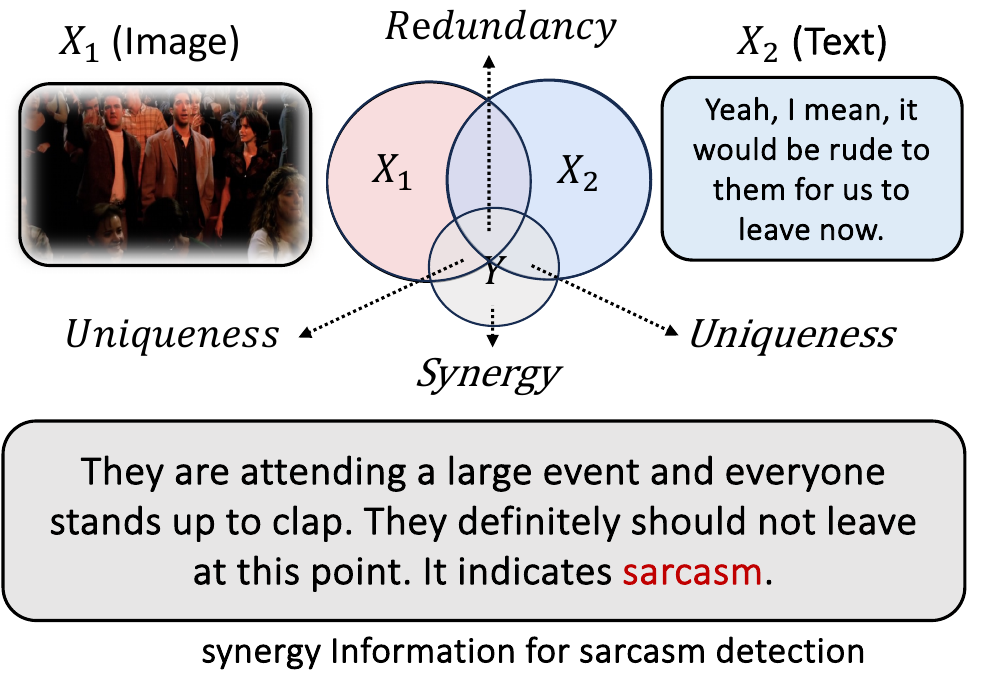}
    \vspace{-1.6mm}
    \caption{\textbf{Case study on synergy interaction}. Existing multimodal models struggle to learn the situation when both text and image modalities alone do not indicate sarcasm, but sarcasm arises due to the synergistic information between modalities when fused together.}
    \label{fig:synergy-case-study}
    \vspace{-2mm}
\end{figure}

To better understand why models struggle with synergy-type interactions, we provide a case study in Figure~\ref{fig:synergy-case-study} that highlights such failure. In this example, both the visual input (\textit{people watching a show and clapping}) and the language input (\textit{they think they should not leave}) lack clear signals of sarcasm individually. However, when combined, the synergized information (\textit{where "them" refers to a band or show and "now" refers to the beginning or ending point of that}) reveals an evident sarcastic intent that is not present in the original visual or language cues. Despite large-scale pretraining, multimodal models struggle to capture such complex interactions between modalities accurately.

\subsubsection*{RQ2. How do expert models perform on corresponding multimodal interaction data?}
While a single large multimodal model may struggle, \model\ leverages specialized expert models to handle each type of interaction. As shown in Table \ref{tab:expert-train-result}, these expert models for redundancy, uniqueness, and synergy outperform test data points with their corresponding interaction types. Notably, expert models for synergy and redundancy show the most significant improvements in \mmsd: Qwen2-0.5B gains over 30 F1 points on synergy, and ALBEF improves by around 8 F1 points on redundancy. In contrast, expert models for uniqueness exhibit almost no change across different model settings. This could be because data points with unique interactions are more prevalent in the dataset compared to those with redundancy or synergy (data points with uniqueness account for around 61\%). As a result, baseline models tend to focus on learning these features during training, leading to similar performance with expert models.

\subsubsection*{RQ3. How small can expert models be?}
\label{scaling-law}
\begin{figure}[ht]
    \centering
    \vspace{-2mm}
    \includegraphics[scale=0.23]{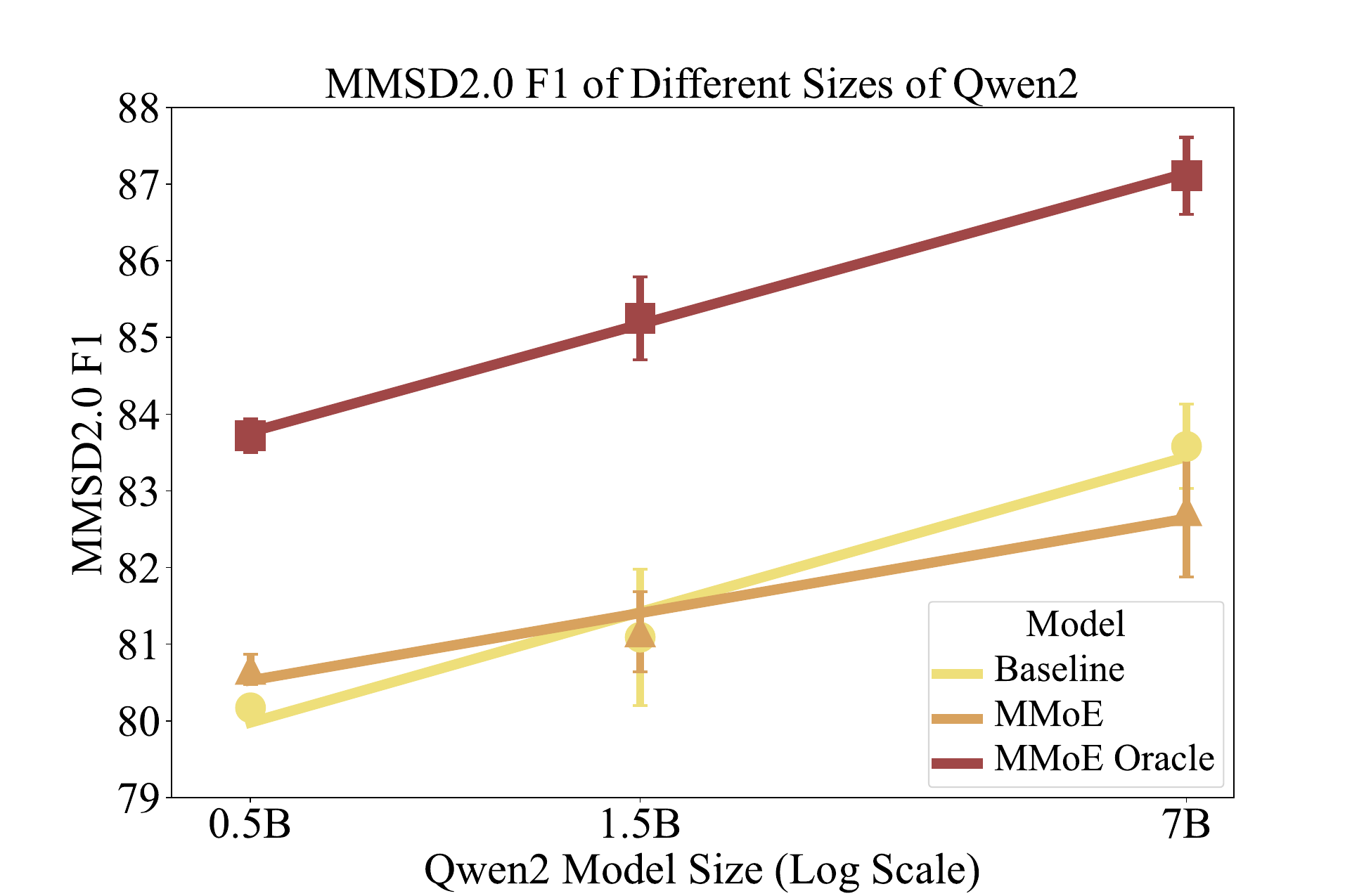}
    \vspace{-2mm}
    \caption{\textbf{\model gains better improvement on smaller models}. \textit{\model Oracle} means that the model fusion process is based on categorized test datapoints with state-of-the-art unimodal models.}
    \label{fig:scaling-law}
    \vspace{-2mm}
\end{figure}

It is well established that neural networks, given enough parameters, are universal function approximators. Therefore, sufficiently large multimodal models should eventually be capable of learning all interaction types. However, we hypothesize that expert models can be smaller and benefit more from \model. To explore the scaling law of \model, we conducted an empirical study using Qwen2 models of different sizes (0.5B, 1.5B, and 7B). We observed a linear relationship between model size and performance score when plotted on a log-scale x-axis, as shown in Figure \ref{fig:scaling-law}. 

\begin{table}[t!]
\centering
\small 
\vspace{2mm}
\begin{tabular}{@{}p{1.9cm}lccc@{}}
    \toprule[1.1pt]
    \textbf{Model} & \textbf{Training} & \textbf{R} & \textbf{U} & \textbf{S} \\
    \midrule
    \multirow{2}{*}{ALBEF} & w/o expert train & 88.70 & 76.02 & 24.39\\
     & w/ expert train & \underline{96.66} & \underline{76.33} & \underline{28.95} \\
    \midrule
    \multirow{2}{*}{BLIP2} & w/o expert train & 96.89 & \underline{80.16} & 20.56 \\
     & w/ expert train & \underline{99.10} & \underline{80.16} & \underline{48.98} \\
    \midrule
    \multirow{2}{*}{Qwen2-0.5B} & w/o expert train & 93.71 & 76.14 & 21.43 \\
     & w/ expert train & \underline{96.54} & \underline{76.16} & \underline{53.66} \\
    \bottomrule[1.1pt]
\end{tabular}
\vspace{-2mm}
\caption{\textbf{Performance of expert models on \mmsd.} Expert training based on the corresponding interaction type improves the model's ability to predict test data points of the same type.}
\vspace{-2mm}
\label{tab:expert-train-result}
\end{table}

When applying \model\ to the 7B model, its performance worsens compared to the single-model baseline. However, as the model size decreases, the benefits of \model\ become increasingly significant. This scaling law suggests that \model\ is more effective with smaller expert models with worse single model performance, which makes sense since smaller models typically struggle to handle multiple interaction types, and specialized expert models can address this limitation more effectively by training on data with specific types of interactions.

Additionally, we also include the \textit{oracle} performance of \model\ when using an oracle router for classifying interaction types in Figure \ref{fig:scaling-law}. With such a router, each data point is always directed to the appropriate expert model for inference. In this setting, the mixture of experts achieves significantly higher performance compared to baseline models and shows a steeper slope when scaling to larger models. This finding suggests that the primary bottleneck of \model\ lies in training an accurate router to route data to the correct expert model for each interaction type. Moreover, it highlights that a model's imbalanced ability to handle different types of multimodal interactions persists regardless of its size or baseline performance.

\subsection*{RQ4. Is the improvement of \model primarily driven by model ensembling?} 
We investigate whether the performance gains in \model are primarily driven by our proposed multimodal interaction-driven data categorization (into redundancy, uniqueness, and synergy) instead of simple multiple model ensembling. To test this hypothesis, we conduct an ablation study using the \urfunny dataset. In this experiment, we kept the number of training data points for each expert model unchanged but replaced the corresponding data points with the ones randomly sampled from the dataset. To eliminate any potential influence introduced by the different fusion methods during inference, we calculate the cross-entropy loss from the three expert models with the smallest values and averaged the score on the whole dataset to assess the upper-bound performance of the mixtures of experts. The metric is defined as:
\begin{align}
    \text{CE}_{moe} = \frac{1}{N}\sum_{i=1}^{N} \min\limits_{y \in \{y_r, y_u, y_s\}} \text{CE}(y, y^*)
\end{align}

\noindent where $N$ represents the total number of the dataset, $y_r$, $y_u$, and $y_s$ represents the logits from expert models and $y^*$ represents the ground-truth labels. We show that for our multimodal interaction-based categorization, the cross-entropy loss is 0.5853 while for random sampling categorization, the cross-entropy loss is 0.6942 (18.59\% increase compared with our proposed categorization). Additionally, the original single model baseline has a loss of 0.8070. It indicates that our methods help build better models for the whole dataset.

\subsection*{RQ5. What do unimodal predictions look like?}
\label{unimodal}

The quality of unimodal partial labels is crucial for accurate data categorization, as these labels directly influence the categorization process. As discussed in Section \S \ref{sec:classify-interaction}, we utilize state-of-the-art models to generate unimodal predictions for the training set. Table \ref{tab:expert-train-result} demonstrates that across all datasets—including \sarcasm, \urfunny, and \mmsd—there is a clear bias toward the text modality. Text-based predictions are 16 points more accurate than those based on visual information. Moreover, predictions from the visual modality exhibit significantly lower confidence compared to those from the text-based modality, indicating that the visual side offers few reliable features for model predictions.

\section{Ablation Study}
\label{sec:ablation-study}

We conduct ablation studies on technical details in stages of categorizing and inference.

\begin{table}[t!]
\centering
\small 

\vspace{1mm}
\begin{tabular}{@{}p{0.9cm}cccccc@{}}
    \toprule[1.1pt]
    \multirow{2}{*}{\textbf{Dataset}} & \multicolumn{3}{c}{\textbf{Text Modality}} & \multicolumn{3}{c}{\textbf{Vision Modality}} \\
    \cmidrule(lr){2-4} \cmidrule(lr){5-7}
    & \textbf{Acc} & \textbf{F1} & \textbf{Conf} & \textbf{Acc} & \textbf{F1} & \textbf{Conf} \\
    \midrule
    \sarcasm & \underline{66.96} & \underline{65.45} & 0.93 & 50.87 & 64.72 & 0.57\\ 
    \urfunny & \underline{67.87} & 61.20 & 0.97 & 50.39 & \underline{62.27} & 0.48\\ 
    \mmsd & \underline{66.88} & 55.74 & 0.97 & 49.58 & \underline{60.39} & 0.63 \\ 
    \bottomrule[1.1pt]
    \end{tabular}
    \vspace{0mm}
    \caption{\textbf{Quality and confidence of unimodal prediction}. \textit{Conf} refers to the confidence of a prediction, calculated as the average of the maximum logits for the tokens "Yes" and "No" from the model's final output logits over the entire vocabulary.}
    \label{tab:unimodal-label}
\end{table}

\begin{table}[t!]
\centering
\small 
\vspace{1mm}
\begin{tabular}{@{}p{2.8cm}ccccc@{}}
    \toprule[1.1pt]
    \textbf{Fusion Method} & \textbf{\sarcasm} & \textbf{\urfunny} & \textbf{\mmsd}  \\
    \midrule
    Baseline & 47.90 & 68.87 & 78.87 \\
    \midrule
    Average fusion & 47.16 & 69.17 & 80.34 \\
    Maximum fusion & 47.84 & 69.55 & 80.70 \\ 
    Weighted fusion & 48.86 & 69.39 & 80.25 \\ 
    Model-based fusion & \underline{48.97} & \underline{70.20} & \underline{80.71} \\ 
    \midrule     
    Oracle fusion & \textbf{56.89} & \textbf{73.36} & \textbf{82.73} \\
    \bottomrule[1.1pt]
    \end{tabular}
    \vspace{0mm}
    \caption{\textbf{Ablation study on various fusion methods on ALBEF}. \textit{Baseline} indicates the single model performance of ALBEF without fusion. \textit{Oracle} refers to fusion performed on the test set that has been categorized using the same method applied to the training data.}
    \vspace{-2mm}
    \label{tab:fusion-method-ablation-study}
\end{table}

\subsection{Ablation study on data categorization}  
\label{expert-model-train-ablate}
We find that having \textit{high-quality} categorized data is crucial for effective expert model training. Often, unimodal information alone doesn’t provide enough useful input for predictions, leading expert models to train on noisy data. This issue is particularly pronounced with vision-based predictions, as discussed in Section \S\ref{unimodal}. To ensure expert models are trained on data that reflects unique interaction type, we filter out any data points where $|p(\text{Yes}) - p(\text{No})| < \delta$, with $\delta$ being a threshold indicating the confidence of the prediction. In experiments with BLIP2 on \urfunny, when $\delta=0$, meaning all training data is used, we achieve a model-based fusion result of 73.64 F1 score. With $\delta=0.1$, partial data points are included in the training, and the F1 score improves to 74.65. However, when we increase $\delta$ to 0.15, the F1 score drops to 73.99, likely due to the reduction in training data. Therefore, we keep $\delta=0.1$ for expert data filtering in our main experiments.

Another technique for expert model training is to rebalance the unimodal predictions of the data to prevent highly imbalanced label distributions after data categorization. Rebalancing helps avoid training collapse in expert models, especially synergy expert models where the training data is few. Further details on data filtering and label rebalancing can be found in Appendix \ref{data-categorization}.

\subsection{Ablation study on expert model fusion} 
We also explore how different fusion strategies for combining multiple expert models impact performance. As mentioned in Section \S\ref{scaling-law}, fusion methods play a significant role during inference, suggesting that each expert model focuses on different aspects of multimodal information, and mixing them up simply cannot take full use of their prediction ability. The common fusion methods we consider include: (1) \textit{Average Fusion}: where we simply average the softmaxed logits from the expert models to produce the final result. (2) \textit{Maximum Fusion}: where we select the highest logits from all the expert models as the final prediction. (3) \textit{Weighted Fusion}: for each dataset, we assign a fixed weight to each expert model, with the weights reflecting the proportion of each interaction type within the whole dataset. (4) \textit{Model-based Fusion}: where we use a BLIP2-based classifier trained to distinguish between redundancy, uniqueness, and synergy. This classifier dynamically adjusts the weights for each expert model for each data point accordingly. Based on Table~\ref{tab:fusion-method-ablation-study}, we find that model-based fusion generally provides the most significant improvement compared with other model-free methods and single-model baseline. However, even a simple model-free fusion can bring improvement on \urfunny and \mmsd datasets, indicating the robustness of our methods.

\section{Conclusion}
\label{sec:conclusion}

This paper proposes a method to enhance multimodal models with a new \textbf{M}ultimodal \textbf{M}ixtures \textbf{o}f \textbf{E}xperts structure (\model). The key idea is to train separate expert models each tailored to learn a specific type of multimodal interaction (including redundancy, uniqueness, and synergy), which overcomes significant shortcomings of existing models when diverse types of interactions are simultaneously present. Categorizing data points into their interactions enables the fusion of expert models during inference, which provides improvement to performance. \model\ also presents improved transparency of the multimodal modeling process. 

\vspace{-1mm}
\section*{Limitations} While we present a first step towards classifying and learning multimodal interactions, our categorization is still at a rather coarse level with only three interactions. Future work should investigate sub-categorizations of interactions, such as different types of synergy between modalities. This can be used to learn mixtures of interactions at a more fine-grained feature level. Furthermore, even approximate classification of interactions can lead to improved performance, so we expect future improvements in quantifying interactions to further improve \model. Future work can also investigate how to better combine multiple interactions in a compositional, multi-step manner to learn more complex higher-order interactions between modalities. Finally, we only consider modalities that have high-quality unimodal encoders like language and vision, future work can extend this direction to novel modalities such as sensors and medical data where unimodal models might have to be learned end-to-end with the multimodal interactions.

\vspace{-1mm}
\section*{Ethics Statement}

Multimodal AI systems can revolutionize many areas involving sensing and prediction such as in multimedia, healthcare, affective computing, and education, but there are also potential negative impacts involving monitoring and tracking humans and their states. For example, emotion detection models can be used inappropriately and invade personal privacy. Careful deployment to mitigate potential risks would be important.

\vspace{-1mm}
\section*{Acknowledgement}

This material is based upon work partially supported by National Science Foundation awards 1722822 and 1750439, National Institutes of Health awards R01MH125740, R01MH132225, R01MH096951 and R21MH130767, and Meta.
PPL is supported in part by a Siebel Scholarship and a Waibel Presidential Fellowship.
RS is supported in part by ONR grant N000142312368 and DARPA FA87502321015.
Any opinions, findings, conclusions, or recommendations expressed in this material are those of the author(s) and do not necessarily reflect the views of the sponsors, and no official endorsement should be inferred.
We thank A100 and H100 GPU support from NetMind.AI\footnote{https://netmind.ai/home} and NVIDIA.

\bibliography{custom}

\clearpage
\appendix

\section{Asset}
\label{asset}
In this section, we list all the necessary information for our use of models and data. In our paper, we use \sarcasm~\citep{castro2019towards}, \urfunny~\citep{hasan2019ur}, \mmsd~\citep{qin-etal-2023-mmsd2} and \mmsarcasm~\citep{cai-etal-2019-multi} for our dataset usage. We use ALBEF~\cite{li2021align}, BLIP2-OPT-2.7B~\cite{li2023blip}, Qwen2-0.5B-Instruct~\citep{qwen2}, Qwen2-1.5B-Instruct, Qwen2-7B-Instruct, Qwen2-72B-Instruct, CogVLM2-LLaMA3-chat-19B~\citep{wang2023cogvlm} for our model usage. We show the required information about them and how we follow their requirements when using them.

\subsection{Model and Data License}
\textbf{ALBEF} (\href{https://github.com/salesforce/ALBEF/tree/main}{download link})\\
License: BSD 3-Clause "New" or "Revised" \\ 
\textbf{BLIP2-OPT-2.7B} (\href{https://github.com/salesforce/LAVIS/tree/main/projects/blip2}{download link})\\ 
License: BSD 3-Clause "New" or "Revised" \\
\textbf{Qwen2-0.5B-Instruct} (\href{https://huggingface.co/Qwen/Qwen2-0.5B-Instruct}{download link})\\
License: Apache 2.0 \\ 
\textbf{Qwen2-1.5B-Instruct} (\href{https://huggingface.co/Qwen/Qwen2-1.5B-Instruct}{download link})\\
License: Apache 2.0 \\
\textbf{Qwen2-7B-Instruct} (\href{https://huggingface.co/Qwen/Qwen2-7B-Instruct}{download link})\\
License: Apache 2.0 \\
\textbf{Qwen2-72B-Instruct} (\href{https://huggingface.co/Qwen/Qwen2-72B-Instruct}{download link})\\
License: Apache 2.0 \\
\textbf{CogVLM2-LLaMA3-chat-19B} (\href{https://huggingface.co/THUDM/cogvlm2-llama3-chat-19B}{download link})\\
License: Apache 2.0  

\subsection{Data License}
\textbf{\sarcasm} (\href{https://github.com/soujanyaporia/MUStARD}{download link}) \\ 
License: MIT \\ 
\textbf{\mmsarcasm} (\href{https://github.com/wrk226/pytorch-multimodal_sarcasm_detection}{download link}) \\
License: Open source, license not specified \\
\textbf{\mmsd} (\href{https://github.com/JoeYing1019/MMSD2.0}{download link})\\ 
License: Open source, license not specified  \\ 
\textbf{\urfunny} (\href{https://github.com/ROC-HCI/UR-FUNNY}{download link})\\
License: Open source, license not specified

\subsection{Model and Data Use}
\paragraph{Personally identifiable information} All of the used datasets in this paper are derived from public sources. Therefore, there is no exposure of any personally identifiable information that requires informed consent from those individuals. The used dataset relates to people insofar as it draws text from public sources that relate to people, or people created, obeying related licenses.
\paragraph{Offensive content claim} All the used datasets including \sarcasm and \mmsarcasm are already public and widely used.  While these datasets may contain instances of offensive content, our work does not aim to generate or amplify such content. Instead, we employ these datasets to study and understand the nature of sarcasm in text. Our use of these datasets follows ethical guidelines, and we do not endorse or support any offensive material contained within them. Moreover, we have implemented measures to mitigate the propagation of offensive content within our research.

\section{Additional Experimental Results}
\label{additional-exp-result}
Besides the models listed in our main sections, we test under more experimental settings with more models. Additionally, we include more baselines for comparison. We also include metrics of precision and recall besides F1 and accuracy that have already been included in the main section. Table \ref{tab:full-table} shows comprehensive experimental results on all the settings that we run and compare. 

\subsection{Model Details}
\paragraph{Model Name} To simplify the terminology in our paper, we use short names for our models. For instance, when we mention BLIP2, we are referring to BLIP2-OPT-2.7B. Similarly, when we refer to Qwen2-0.5B/1.5B/7B/72B, this corresponds to Qwen-2-0.5-0.5B/1.5B/7B/72B-Instruct. Lastly, CogVLM2 refers to CogVLM2-LLaMA3-chat-19B.

\paragraph{Model Size} ALBEF consists of a BERT base model with 123.7 million parameters and a ViT-B/16 with 85.8 million parameters, bringing the total to 209.5 million. BLIP2, on the other hand, includes a 2.7 billion-parameter OPT model, a Q-Former, and a ViT. Since the Q-Former and ViT are relatively small compared to OPT, the total size of BLIP2 is approximately 2.7 billion parameters. For the Qwen models, the number of parameters corresponds to the model names: 0.5B, 1.5B, 7B, and 72B. Lastly, CogVLM2 includes a ViT-style vision encoder and a 19 billion-parameter LLaMA3-chat checkpoint. Since the vision encoder and projection parameters are much smaller than LLaMA3-chat, the total size of CogVLM2 is around 19 billion parameters.

\section{Dataset Details}
\label{dataset-info}
\begin{table}
\centering
\caption{\textbf{Statistical information for 4 multimodal datasets that we use in our experiments.} $\dagger$ indicates that the validation split is not provided in the original dataset and is conducted by randomly sampling from training data by ourselves.}
\begin{tabular}{@{}p{2.8cm}lll@{}}
    \toprule[1.1pt]
    Dataset       & \#Train & \#Valid & \#Test \\
    \midrule
    \sarcasm & 251$\dagger$ & 83$\dagger$ & 356 \\ 
    \mmsarcasm & 29,040 & 2,410 & 2,409 \\
    \mmsd & 19,816 & 2,410 & 2,409 \\
    \urfunny & 7,614 & 980 & 992 \\
    \bottomrule[1.1pt]
    \end{tabular}
    \label{tab:data_stats}
\end{table}

Statistical information for the splits of 4 multimodal datasets included in our experiments is shown in Table \ref{tab:data_stats}.
We introduce the basic information for each dataset in the following.

\paragraph{\sarcasm} contains 690 videos with evenly balanced sarcasm and non-sarcasm labeled points. This dataset is based on English and mainly collected from TV show clips including \textit{Friends}, \textit{The Big Bang Theory}, \textit{The Golden Girls}, and \textit{Sarcasmaholics Anonymous}. Its domain mainly covers daily conversation. The annotation is conducted by two graduate students in two steps: annotating \textit{The Big Bang Theory} first and annotating the remaining ones. Additionally, we use the speaker-independent training and testing splits to make sure that there is no overlap between speakers in the training and testing sets to avoid potential bias.

\paragraph{\mmsarcasm} collects English tweets containing a picture and some special hashtag (e.g., \#sarcasm, etc.) as positive examples (i.e. sarcastic) and collects English tweets with images but without such hashtags as negative examples (i.e. not sarcastic). Furthermore, it excludes tweets with keywords like sarcasm, sarcasm, irony, and irony. Moreover, it discards tweets containing URLs to avoid introducing additional information and discards tweets with words that frequently co-occur with sarcastic tweets and thus may express sarcasm, for instance, jokes, humor, and engagement.

\paragraph{\mmsd} is a polished version of \mmsarcasm.  It removes the spurious cues (e.g., sarcasm word) from the text in the \mmsarcasm, which encourages the model to truly capture the relationship across different modalities rather than just memorize the spurious correlation. Additionally, it re-annotates the unreasonable data in \mmsarcasm. Therefore, the text information in \mmsd is slightly different from \mmsarcasm and part of the labels are re-annotated.

\paragraph{\urfunny} is a collection of 1866 TED talks, as well as their transcripts, including 1,741 speakers and 417 topics that include speakers from different backgrounds and nationalities and topics from scientific discoveries to everyday ordinary events. The laughter markup is used to filter out 8,257 humorous punchlines from the transcripts. The last sentence is assumed a punchline and similar to the positive instances, the context is chosen.

\section{Dataset Preprocessing Details}
\label{data-preprocess}
Different multimodal datasets require different preprocessing methods before conducting model training. We include the details of our preprocessing in this section.

\paragraph{\mmsarcasm and \mmsd} We are only able to extract a total of 24635 images from the released dataset and thus filtered the dataset by the existence of corresponding image IDs. The sizes of validation and test sets are unaffected, while the number of training instances drops to 19,816. 

\paragraph{\sarcasm and \urfunny} There are no existing keyframes in the original dataset. We had to split the videos into frames for use in our image-text models. Typically, we find that key frames matter a lot for the multimodal prediction. Therefore, we used FFmpeg, where we used 1 frame per second to split into frames. Out of the frames extracted per video, we choose the most representative frame by conducting facial expression recognition by DeepFace~\cite{deepface} and selecting the frame with the highest emotion intensity score. We thus created the image modality off on the original video dataset. 

\section{Image Description Details}
\label{appendix:image-description}

To allow the applicability of our method to pure text-based LLMs, we convert each image into detailed descriptions that include task-related information. We include the details about the process of using CogVLM2-LLaMA3-chat-19B to achieve this in Table \ref{tab:image-description-prompt1}, \ref{tab:image-description-prompt2}, and \ref{tab:image-description-prompt3}.

\section{Data Categorization Details}
\label{appendix:unimodal-label}

To achieve the dataset categorization based on three types of multimodal interaction including redundancy, uniqueness, and synergy. We need to finish this in multiple steps: (1) vision-based prediction collection (2) text-based prediction collection (3) multimodal data categorization. In the following section, we include the technical details for each of them.

\subsection{Vision-based Prediction Collection}
\label{appendix:vision-only-label}
We utilize CogVLM2-LLaMA3-chat-19B as our base model vision-only prediction collection. Typically, even though CogVLM2-LLaMA3-chat-19B is a multimodal model, we only include image-side information and only add task-related queries like "Is the image sarcastic or not?" as the input to make sure the model does not utilize text-side information from the multimodal dataset to do the prediction. We conduct few-shot prompting on the train and validation split of all multimodal models. Since \mmsarcasm and \mmsd share the same set of images and conduct the same multimodal prediction task, we show three prompts that are used for 4 multimodal datasets in Table \ref{tab:mustard-vision-prompt}, \ref{tab:mmsd-vision-prompt}, and \ref{tab:urfunny-vision-prompt}.

\subsection{Text-based Prediction Collection}
\label{appendix:text-only-label}

We utilize {Qwen2-72B-Instruct as our base model for text-only prediction. Typically, we only include text-side information and task-related queries like "Is this image sarcastic or not?" as the input to make multimodal predictions. Even though \mmsarcasm and \mmsd do not share the same setting, most of their data is similar. Therefore, we utilize the same prompts for them. We show three few-shot prompts that are used for 4 multimodal datasets in Table \ref{tab:mustard-text-prompt}, \ref{tab:sarcasm-text-prompt}, \ref{tab:urfunny-text-prompt}.

\subsection{Multimodal Data Categorization}
\label{data-categorization}
After collecting unimodal predictions for all multimodal datasets, we conduct our algorithm for categorizing each data point into different multimodal interaction types (redundancy, uniqueness, and synergy) to make sure our categorized data is suitable for training. Typically, to achieve more robust and effective data categorization, we design filtering and rebalancing stages as part of categorization.

\paragraph{Filtering} After collecting prediction logits with the model \texttt{CogVLM2-LLaMA3-chat-19B}, we collect the output logits for predictions, which reflect the model's confidence for “Yes” or “No” responses. To finalize the predictions in multimodal tasks, we apply a softmax operation on these logits to convert them into probabilities. We observed that relying only on vision-related information might lead to inaccurate or uncertain predictions. Therefore, to enhance the reliability of the training data, we remove any data points where the prediction confidence is below 0.55. These low-confidence predictions are seen as lacking clear patterns and could introduce noise into the training process. By filtering out these training data points, we aim to improve the overall quality and accuracy of the model's predictions.

\paragraph{Rebalancing} Filtering guarantees a high-quality set of pseudo labels. However, the bias from a single modality might cause significant bias in the prediction. The extremely imbalanced distribution of the pseudo labels might lead to the model overfitting to the majority class. To address this issue, we rebalance the dataset by undersampling the majority class. We rank the probability of the prediction from high to low. If the minority of the class is more than 20\% of the overall dataset number, we keep the prediction as it is. Otherwise, we consider it as an extremely unbalanced case and make sure that the minority class is at least 20\% of the overall dataset to avoid extreme imbalance in the dataset. This helps us avoid expert models (including redundancy, uniqueness, and synergy) overfitting to the majority class and ensures that the model is trained on a balanced dataset.

\paragraph{Categorizing} Upon finishing the filtering and rebalancing stage, we have groups of high-quality and balanced unimodal predictions. Therefore, combining it with the ground-truth labels, we categorize the dataset into redundancy, uniqueness, and synergy separately on train, validation, and test splits. The algorithm that is used to conduct the categorization is below.

\begin{algorithm} \caption{Multimodal Categorization} \label{alg
} \begin{algorithmic}[1] \Require Text-based label $y_1$, Vision-based label $y_2$, Ground-truth label $y^{}$ \Ensure Interaction category: $R$, $U$, or $S$ \If {$y_1 = y_2$ \textbf{and} $y_1 = y^{}$} \State \Return $R$ \ElsIf {$y_1 = y_2$} \State \Return $S$ \ElsIf {$y_1 = y^{}$ \textbf{or} $y_2 = y^{}$} \State \Return $U$ \Else \State \Return $S$ \EndIf \end{algorithmic} \end{algorithm}

\vspace{-2mm}
\section{Expert Model Training Details}
\label{expert-model-train-detail}
To improve expert training, we find that instead of starting from the initial pre-trained model checkpoint, it's more effective to initialize the expert training phase using fine-tuned baseline models. This approach leads to faster training and better overall results. The reasoning behind this decision is that continuing training from an already fine-tuned model allows the model to build on its learned features while still maintaining strong performance across the entire dataset. Preserving this capability is essential during inference because the fusion process might assign incorrect nodes to the wrong expert models, and maintaining some general competency helps mitigate such errors from the fusion model and achieve better general performance.

\section{Fusion Model Training Details}
\label{fusion-train}
To conduct a model-based fusion, we need to train a fusion model. We use BLIP2 for classifying multimodal interactions and focus on three key categories: redundancy, uniqueness, and synergy. However, these categories are often imbalanced in datasets such as \sarcasm, \urfunny, and \mmsd, with certain types being underrepresented. To address this imbalance problem, we adopt focal loss as the optimization target:
\[
\text{FL}(p_t) = -\alpha_t (1 - p_t)^\gamma \log(p_t)
\]
where we set $\alpha=1$ and $\gamma=2$.

\section{Experimental Details}
\label{appendix:exp-details}
We include all the technical details of our experiments including computational requirements and hyper-parameter settings.

\subsection{Computational Costs}
We utilize 5$\times$A6000 or 1$\times$A100 to run baseline experiments. Expert model training approximately requires 1.5 times longer than baseline training since we need to train redundancy, uniqueness, and synergy models separately. The fusion model training includes a similar training configuration with baselines but just trains under a 3-class classification.

\subsection{Hyper-parameter Settings}
We use different sets of hyperparameters for the various training settings, including baseline training, expert model training, and fusion model training. We do not perform hyperparameter searches but instead tune the parameters based on the validation set. For LoRA-based fine-tuning, we generally set the maximum sequence length to 512, rank to 16, scaling factor to 32, and dropout rate to 0.05.

For baseline training, the specific hyperparameters are as follows:

\begin{itemize}
    \item \textbf{For \sarcasm:} We use 10 epochs, a learning rate of 4e-5, evaluation steps every 100 iterations, and a batch size of 40 for ALBEF and BLIP2. For Qwen2 models, the number of epochs is also set to 10, evaluating at the end of each epoch, and the batch size is 1.
    
    \item \textbf{For \urfunny:} We use 4 epochs, a learning rate of 5e-5, evaluation steps every 100 iterations, and a batch size of 10 for ALBEF, BLIP2. We use a batch size of 1 for Qwen2, evaluating at the end of each epoch, and the number of epochs is also set to 10.
    
    \item \textbf{For \mmsd:} We use 4 epochs, a learning rate of 5e-5, evaluation steps every 100 iterations, and a batch size of 10 for ALBEF, BLIP2. We use 5 epochs and a batch size of 1, also evaluating at the end of each epoch, for Qwen2.
\end{itemize}

For expert model training, we increase the number of epochs to 10, while keeping the other hyperparameters unchanged, to ensure sufficient training.

For fusion model training, the hyperparameters vary across datasets when training BLIP2 on them:

\begin{itemize}
    \item \textbf{For \sarcasm:} We use 50 epochs, a learning rate of 1e-4, evaluation steps every 20 iterations, and a batch size of 50.
    
    \item \textbf{For \urfunny:} We use 50 epochs, a learning rate of 1e-4, evaluation steps every 70 iterations, and a batch size of 50.
    
    \item \textbf{For \mmsd:} We use 20 epochs, a learning rate of 1e-4, evaluation steps every 200 iterations, and a batch size of 50.
\end{itemize}

\subsection{Model Selection Details}
In our experiments, which include baseline training, expert model training, and fusion model training, we consistently use the F1 score on the validation set as the metric for model selection. For baseline training, we select the model checkpoint with the highest F1 score on the entire development set. During expert model training, we choose the best expert model checkpoint based on the highest F1 score on the specific subset of the development set that corresponds to the relevant type of multimodal interaction. For fusion model training, we select the model that has the highest 3-class F1 score.

\subsection{Evaluation Details}

We used the metrics module from the \texttt{scikit-learn} package for evaluating our prediction tasks. Since our tasks are binary prediction tasks, we chose the binary averaging strategy for precision, recall, and f1. Additional details can be found in the \texttt{scikit-learn} documentation for the metrics module.

\subsection{Experimental Statistics}
All the available results are based on three different random seeds, with both the mean and standard deviation reported. Typically, F1 results where adding \model leads to a statistically significant change (p-value < 0.05) are marked with a $^*$ in Table \ref{tab:full-table}. F1 results have a p-value < 0.1 and are marked with a $^{**}$ in Table \ref{tab:full-table}.

\section{AI Assistance}

We did use ChatGPT as the writing assistant to help us write part of the paper. Additionally, we utilize the power of CodePilot to help us code faster. However, all the AI-generated writing and coding components assisted by AI are manually checked and modified. There is no full AI-generated content in the paper.

\begin{table*}[ht]
\centering
\small 
\caption{\textbf{Comprehensive results on all types of models and different datasets.} The numbers in the table represent the mean values from 3 runs with 3 seeds, with the corresponding standard variance provided. $\dagger$ indicates that the results include information from audio modality while ours does not.}
\begin{tabular}{@{}lp{2cm}p{2cm}p{2cm}p{2cm}@{}}
    \toprule[1.1pt]
    \textbf{Model}       & \textbf{Acc} ($\uparrow$) & \textbf{Precision} ($\uparrow$) & \textbf{Recall} ($\uparrow$) & \textbf{F1} ($\uparrow$) \\
    \midrule
    \midrule
    \multicolumn{5}{c}{\sarcasm} \\ 
    \midrule
    MulT$\dagger$~\citep{tsai2019multimodal} & - & 65.51 & 64.78 & 64.49  \\
    LMF$\dagger$~\citep{liu2018efficient} & - & 70.46 & 70.34 & 69.92 \\
    LF-DNN-v2$\dagger$~\citep{ding2022multimodal} & - & 65.95 & 63.88 & 62.30 \\
    LMF~\citep{liu2018efficient} & - & 70.73 & 70.90 & 70.68 \\ 
    LF-DNN-v1$\dagger$~\citep{ding2022multimodal} & - & 71.55 & 71.52 & 70.99  \\
    \midrule 
    ALBEF & 54.49$_{\pm 3.13}$ & 47.08$_{\pm 3.03}$ & 50.22$_{\pm 3.62}$ & 48.51$_{\pm 2.21}$ \\ 
    ALBEF+\model & 54.49$_{\pm 2.85}$ & 47.36$_{\pm 2.72}$ & 57.68$_{\pm 4.76}$ & 51.95$_{\pm 2.81}$ \\ 
    \midrule
    BLIP2 & 53.75$_{\pm 9.33}$ & 48.46$_{\pm 4.94}$ & 90.13$_{\pm 9.21}$ & 62.65$_{\pm 2.67}$ \\ 
    BLIP2+\model & 59.18$_{\pm 2.11}$ & 51.26$_{\pm 1.38}$ & 87.94$_{\pm 6.11}$ & 64.74$_{\pm 2.49}$ \\ 
    \midrule
    Qwen2-0.5B & 54.59$_{\pm 4.35}$ & 48.35$_{\pm 3.31}$ & 74.12$_{\pm 9.40}$ & 58.17$_{\pm 0.86}$ \\ 
    Qwen2-0.5B+MMoE & 49.06$_{\pm 3.00}$ & 45.16$_{\pm 1.39}$ & 88.60$_{\pm 3.97}$ & 59.77$_{\pm 0.35}^{**}$ \\ 
    \midrule 
    Qwen2-1.5B& 64.79$_{\pm 4.11}$ & 56.45$_{\pm 3.53}$ & 78.73$_{\pm 13.18}$ & 65.38$_{\pm 5.16}$ \\ 
    Qwen2-1.5B+MMoE & 70.69$_{\pm 3.28}$ & 60.86$_{\pm 3.58}$ & 89.47$_{\pm 3.29}$ & 72.34$_{\pm 1.50}$ \\ 

    \midrule 
    Qwen2-7B& 72.75$_{\pm 0.74}$ & 63.27$_{\pm 1.56}$ & 86.62$_{\pm 4.38}$ & 72.91$_{\pm 0.74}$ \\ 
    Qwen2-7B+MMoE & 70.41$_{\pm 3.23}$ & 60.64$_{\pm 3.57}$ & 89.04$_{\pm 3.38}$ & 71.78$_{\pm 1.47}$ \\ 
    \midrule 
    \midrule
    \multicolumn{5}{c}{\urfunny} \\
    \midrule 
    MulT$\dagger$~\citep{tsai2019multimodal} & 66.65 & - & - & -  \\
    FDMER~\citep{yang2022disentangled} & 70.43 & - & - & - \\
    MMIM+SuCI$\dagger$~\citep{yang2024towards} & 70.92 & - & - & - \\
    FDMER$\dagger$~\citep{yang2022disentangled} & 71.87 & - & - & - \\
    \midrule 
    ALBEF & 66.77$_{\pm 0.24}$ & 64.29$_{\pm 1.08}$ & 73.74$_{\pm 2.90}$ & 68.67$_{\pm 0.79}$ \\ 
    ALBEF+MMoE & 67.91$_{\pm 0.27}$ & 65.17$_{\pm 0.30}$ & 75.24$_{\pm 1.53}$ & 69.85$_{\pm 0.52}^{*}$ \\ 
    \midrule 
    BLIP2 & 70.43$_{\pm 0.20}$ & 65.14$_{\pm 0.23}$ & 86.60$_{\pm 1.07}$ & 74.31$_{\pm 0.35}$ \\ 
    BLIP2+MMoE & 71.27$_{\pm 0.30}$ & 66.60$_{\pm 1.23}$ & 84.15$_{\pm 1.95}$ & 74.32$_{\pm 0.36}$ \\ 
    \midrule 
    Qwen2-0.5B & 69.29$_{\pm 0.54}$ & 67.16$_{\pm 1.70}$ & 74.15$_{\pm 1.85}$ & 70.46$_{\pm 0.14}$ \\ 
    Qwen2-0.5B+MMoE & 69.19$_{\pm 0.14}$ & 69.36$_{\pm 0.07}$ & 67.55$_{\pm 0.44}$ & 68.38$_{\pm 0.20}$ \\ 
    \midrule 
    Qwen2-1.5B& 70.43$_{\pm 0.53}$ & 66.03$_{\pm 0.41}$ & 83.13$_{\pm 2.27}$ & 73.51$_{\pm 0.89}$ \\ 
    Qwen2-1.5B+MMoE & 68.25$_{\pm 0.81}$ & 64.40$_{\pm 1.87}$ & 80.07$_{\pm 1.37}$ & 71.34$_{\pm 0.52}$ \\ 
    \midrule 
    Qwen2-7B & 72.41$_{\pm 0.52}$ & 68.14$_{\pm 0.65}$ & 82.93$_{\pm 0.43}$ & 74.80$_{\pm 0.55}$ \\ 
    Qwen2-7B+MMoE & 71.88$_{\pm 0.51}$ & 69.18$_{\pm 0.67}$ & 78.16$_{\pm 2.56}$ & 73.29$_{\pm 0.86}$ \\
    \midrule
    \midrule
    \multicolumn{5}{c}{\mmsd} \\
    \midrule 
    HKE~\citep{liu-etal-2022-towards-multi-modal} & 76.50 & 73.48 & 71.07 & 72.25 \\
    ViT~\citep{dosovitskiy2020image} & 72.02 & 65.26 & 74.83 & 69.72 \\
    DynRT-Net~\citep{tian-etal-2023-dynamic} & 71.40 & 71.80 & 72.17 & 71.34 \\
    Multi-view CLIP~\citep{qin-etal-2023-mmsd2} & 85.64 & 80.33 & 88.24 & 84.10 \\
    ChatGLM2-6B~\citep{du2022glm} & 80.08 & 80.52 & 81.04 & 80.04 \\
    LLaMA2-7B~\citep{touvron2023llama} & 84.68 & 84.40 & 84.94 & 84.53 \\
    LLaVA1.5-7B~\citep{liu2024improved} & 85.18 & 85.89 & 85.20 & 85.11 \\
    LLaVA1.5-7B+DemoRetrieval~\citep{tang-etal-2024-leveraging} & 86.43 & 87.00 & 86.30 & 86.34 \\
    \midrule
    ALBEF & 81.79$_{\pm 0.86}$ & 77.58$_{\pm 1.35}$ & 81.23$_{\pm 1.59}$ & 79.33$_{\pm 0.18}$ \\ 
    ALBEF+\model & 82.30$_{\pm 0.31}$ & 76.24$_{\pm 0.24}$ & 85.57$_{\pm 0.42}$ & 80.63$_{\pm 0.32}^{*}$ \\
    \midrule 
    BLIP2 & 84.75$_{\pm 0.99}$ & 78.08$_{\pm 1.65}$ & 89.78$_{\pm 2.71}$ & 83.52$_{\pm 0.04}$  \\
    BLIP2+\model & 84.82$_{\pm 0.87}$ & 78.87$_{\pm 1.60}$ & 88.49$_{\pm 2.36}$ & 83.38$_{\pm 0.05}$ \\
    \midrule
    Qwen2-0.5B & 81.87$_{\pm 0.81}$ & 75.83$_{\pm 1.47}$ & 85.09$_{\pm 1.59}$ & 80.17$_{\pm 0.14}$ \\ 
    Qwen2-0.5B+MMoE & 82.27$_{\pm 0.64}$ & 76.02$_{\pm 1.37}$ & 85.92$_{\pm 4.07}$ & 80.67$_{\pm 1.55}^{*}$ \\ 
    \midrule 
    Qwen2-1.5B& 83.24$_{\pm 1.32}$ & 78.81$_{\pm 2.33}$ & 83.54$_{\pm 4.51}$ & 81.10$_{\pm 0.94}$ \\ 
    Qwen2-1.5B+MMoE & 82.76$_{\pm 0.63}$ & 76.70$_{\pm 1.28}$ & 86.21$_{\pm 3.37}$ & 81.16$_{\pm 0.76}$ \\  
    \midrule 
    Qwen2-7B& 85.28$_{\pm 1.17}$ & 80.38$_{\pm 0.87}$ & 87.05$_{\pm 2.43}$ & 83.58$_{\pm 1.29}$ \\ 
    Qwen2-7B+MMoE & 84.35$_{\pm 1.05}$ & 78.74$_{\pm 2.65}$ & 87.21$_{\pm 4.34}$ & 82.74$_{\pm 0.43}$ \\  
    \bottomrule[1.1pt]
    \end{tabular}

    \label{tab:full-table}
\end{table*}

\begin{table*}
\centering
\caption{\textbf{Prompt for generating image description of \sarcasm}}
\begin{tabular}{p{0.1\textwidth}p{0.8\textwidth}}
\toprule[1.1pt]
\textbf{Role} & \textbf{Content} \\
\midrule
\multirow{5}{*}{\texttt{System}} 
& \texttt{Describe the image in detail.} \\
& \texttt{If there are people, focus on their emotions, postures, facial expressions, body language, and interactions. Based on this information, infer what event is going on.} \\
& \texttt{If there are no people, analyze the event or scene, considering background elements and overall context to infer what event is going on.} \\
& \texttt{Provide evidence to predict if the situation is humorous.} \\
& \texttt{Ensure the description is between 15 to 100 words.} \\
\bottomrule[1.1pt]
\end{tabular}
\label{tab:image-description-prompt1}
\end{table*}

\begin{table*}
\centering
\caption{\textbf{Prompt for generating image description of \mmsarcasm and \mmsd}}
\begin{tabular}{p{0.1\textwidth}p{0.8\textwidth}}
\toprule[1.1pt]
\textbf{Role} & \textbf{Content} \\
\midrule
\multirow{5}{*}{\texttt{System}} 
& \texttt{Describe the image in detail.} \\
& \texttt{If there are people, focus on their emotions, postures, facial expressions, body language, and interactions. Based on this information, infer what is the event going on.} \\
& \texttt{If there are no people, analyze the event or scene, considering background elements and overall context to infer what is the event going on.} \\
& \texttt{Provide evidence to predict if the situation is sarcastic.} \\
& \texttt{Ensure the description is between 15 to 100 words.} \\
\bottomrule[1.1pt]
\end{tabular}
\label{tab:image-description-prompt2}
\end{table*}

\begin{table*}
\centering
\caption{\textbf{Prompt for generating image description of \urfunny}}
\begin{tabular}{p{0.1\textwidth}p{0.8\textwidth}}
\toprule[1.1pt]
\textbf{Role} & \textbf{Content} \\
\midrule
\multirow{5}{*}{\texttt{System}} 
& \texttt{Describe the image in detail.} \\
& \texttt{If there are people, focus on their emotions, postures, facial expressions, body language, and interactions. Based on this information, infer what is the event going on.} \\
& \texttt{If there are no people, analyze the event or scene, considering background elements and overall context to infer what is the event going on.} \\
& \texttt{Provide evidence to predict if the situation is sarcastic.} \\
& \texttt{Ensure the description is between 15 to 100 words.} \\
\bottomrule[1.1pt]
\end{tabular}
\label{tab:image-description-prompt3}
\end{table*}

\begin{table*}
\centering
\caption{\textbf{Prompt for generating image-only prediction of \sarcasm}}
\begin{tabular}{p{0.1\textwidth}p{0.8\textwidth}}
\toprule[1.1pt]
\textbf{Role} & \textbf{Content} \\
\midrule
\multirow{8}{*}{\texttt{System}} 
& \texttt{Please analyze the image provided for sarcastic or not. The image is a screenshot of a TV show.} \\
& \texttt{If you think the image includes exaggerated emotions (like laughing or looking angry or raising eyebrows) or exaggerated posture (like stretching hands), please answer 'Yes'.} \\
& \texttt{If you think the image shows people discussing serious things and just daily routines, please answer 'No'.} \\
& \texttt{You need to think about what is the potential event going on in the image.} \\
& \texttt{Please make sure that your answer is based on the image itself, not on the context or your knowledge.} \\
& \texttt{There are only two options: 'Yes' or 'No'.} \\
& \texttt{If you are not sure, please provide your best guess and do not say that you are not sure.} \\
& \texttt{You should only make No judgment when you are very sure that the image is not sarcastic. As long as you think potentially it is sarcastic, you should say Yes.} \\
\bottomrule[1.1pt]
\end{tabular}
\label{tab:mustard-vision-prompt}
\end{table*}

\begin{table*}
\centering
\caption{\textbf{Prompt for generating image-only prediction of \mmsarcasm and \mmsd}}
\begin{tabular}{p{0.1\textwidth}p{0.8\textwidth}}
\toprule[1.1pt]
\textbf{Role} & \textbf{Content} \\
\midrule
\multirow{9}{*}{\texttt{System}} 
& \texttt{Please analyze the image provided for sarcastic or not. The image is a screenshot of the image on Twitter. It might include a lot of text, so you need to combine the information of the text in the image.} \\
& \texttt{If you think the image includes exaggerated emotions (like laughing or looking angry or raising eyebrows) or exaggerated posture (like stretching hands), please answer 'Yes'.} \\
& \texttt{If you think the image includes text that is sarcastic or exaggerated, please answer 'Yes'.} \\
& \texttt{If you think the image shows people discussing serious things and just daily routines, please answer 'No'.} \\
& \texttt{You need to think about what is the potential event going on in the image.} \\
& \texttt{Please make sure that your answer is based on the image itself, not on the context or your knowledge.} \\
& \texttt{There are only two options: 'Yes' or 'No'.} \\
& \texttt{If you are not sure, please provide your best guess and do not say that you are not sure.} \\
& \texttt{You should only make No judgment when you are very sure that the text is not sarcastic. As long as you think potentially it is sarcastic, you should say Yes.} \\
\bottomrule[1.1pt]
\end{tabular}
\label{tab:mmsd-vision-prompt}
\end{table*}

\begin{table*}
\centering
\caption{\textbf{Prompt for generating image-only prediction of \urfunny}}
\begin{tabular}{p{0.1\textwidth}p{0.8\textwidth}}
\toprule[1.1pt]
\textbf{Role} & \textbf{Content} \\
\midrule
\multirow{11}{*}{\texttt{System}} 
& \texttt{You are looking at a screenshot of a TED talk. It is part of the talk and it can be a slide or a speaker.} \\
& \texttt{Please analyze the image provided to show whether the image is part of a talk that is showing serious content or trying to show some potentially funny content that can make the audience laugh.} \\
& \texttt{If you are looking at a slide, please think about the content of the slide.} \\
& \texttt{If the slide is showing some very interesting and informal things, we believe the speaker is trying to make some jokes, and please answer 'Yes'.} \\
& \texttt{If the slide is showing some very serious and formal things, we believe the speaker is trying to show some serious content and please answer 'No'.} \\
& \texttt{If you are looking at a speaker, please think about the speaker's facial expression and body language.} \\
& \texttt{If you think the image includes exaggerated emotions or its body language is exaggerated, we believe the speaker is talking about some informal things and please answer 'Yes'.} \\
& \texttt{If you think the speaker in the image looks very serious and formal, they are trying to convey their key points and please answer 'No'.} \\
& \texttt{Please make sure that your answer is based on the image itself, not on the context or your knowledge.} \\
& \texttt{There are only two options: 'Yes' or 'No'.} \\
& \texttt{If you are not sure, please provide your best guess and do not say that you are not sure.} \\
\bottomrule[1.1pt]
\end{tabular}
\label{tab:urfunny-vision-prompt}
\end{table*}

\begin{table*}
\centering
\caption{\textbf{Prompt for generating text-only prediction of \sarcasm}}
\begin{tabular}{p{0.1\textwidth}p{0.8\textwidth}}
\toprule[1.1pt]
\textbf{Role} & \textbf{Content} \\
\midrule
\multirow{7}{*}{\texttt{System}} 
& \texttt{Please analyze the text provided below for sarcasm.} \\
& \texttt{If you think the text includes an exaggerated description or includes strong emotion or its real meaning is not aligned with the original one, please answer 'Yes'.} \\
& \texttt{If you think the text is neutral or its true meaning is not different from its original one, please answer 'No'.} \\
& \texttt{Please make sure that your answer is based on the text itself, not on the context or your knowledge.} \\
& \texttt{There are only two options: 'Yes' or 'No'.} \\
& \texttt{If you are not sure, please provide your best guess and do not say that you are not sure.} \\
& \texttt{You should only make Yes judgment when you are very sure that the text is sarcastic.} \\
\texttt{User} & \texttt{TEXT: Yes yes it is! In Prison!!} \\
\texttt{Assistant} & \texttt{Yes. It expresses the speaker's strong emotion about the situation which indicates that the speaker is sarcastic.} \\
\texttt{User} & \texttt{TEXT: And then and then you clicked it again, she's dressed. She is a businesswoman, she is walking down the street and oh oh oh she's naked.} \\
\texttt{Assistant} & \texttt{No. It is a neutral statement.} \\
\bottomrule[1.1pt]
\end{tabular}
\label{tab:mustard-text-prompt}
\end{table*}

\begin{table*}
\centering
\caption{\textbf{Prompt for generating text-only prediction of \mmsarcasm and \mmsd}}
\begin{tabular}{p{0.1\textwidth}p{0.8\textwidth}}
\toprule[1.1pt]
\textbf{Role} & \textbf{Content} \\
\midrule
\multirow{6}{*}{\texttt{System}} 
& \texttt{Please analyze the text provided below for sarcasm.} \\
& \texttt{If you think the text includes an exaggerated description or its real meaning is not aligned with the original one, please answer 'Yes'.} \\
& \texttt{If you think the text is neutral or its true meaning is not different from its original one, please answer 'No'.} \\
& \texttt{Please make sure that your answer is based on the text itself, not on the context or your knowledge.} \\
& \texttt{There are only two options: 'Yes' or 'No'.} \\
& \texttt{If you are not sure, please provide your best guess and do not say that you are not sure.} \\
\texttt{User} & \texttt{TEXT: because lunch is more interesting than job and even tasty...} \\
\texttt{Assistant} & \texttt{Yes. It expresses the speaker's preference for lunch over the job by using the word 'tasty'.} \\
\texttt{User} & \texttt{TEXT: gameday ready'} \\
\texttt{Assistant} & \texttt{No. It is a neutral statement.} \\
\bottomrule[1.1pt]
\end{tabular}
\label{tab:sarcasm-text-prompt}
\end{table*}

\begin{table*}[ht]
    \centering
    \caption{\textbf{Prompt for generating text-only prediction of \urfunny}}
    \begin{tabular}{p{0.1\textwidth}p{0.8\textwidth}}
    \toprule[1.1pt]
    \textbf{\texttt{Role}} & \textbf{\texttt{Content}} \\
    \hline
    \multirow{7}{*}{\texttt{System}} & \texttt{Please analyze the text provided below for humor or not.} \\
    & \texttt{If you think the text includes an exaggerated description or it is expressing sarcastic meaning, please answer 'Yes'.} \\
    & \texttt{If you think the text is neutral or just common meaning, please answer 'No'.} \\
    & \texttt{Please make sure that your answer is based on the text itself, not on the context or your knowledge.} \\
    & \texttt{There are only two options: 'Yes' or 'No'.} \\
    & \texttt{If you are not sure, please provide your best guess and do not say that you are not sure.} \\
    & \texttt{You should only make No judgment when you are very sure that the text is not funny. As long as you think potentially it is funny, you should say Yes.} \\
    \texttt{User} & \texttt{TEXT: why invite men they are the problem} \\
    \texttt{Assistant} & \texttt{Yes. It expresses that men can be problematic and the speaker is sarcastic to make people laugh.} \\
    \texttt{User} & \texttt{TEXT: we all feel the same things.} \\
    \texttt{Assistant} & \texttt{No. It is a neutral statement.} \\
    \bottomrule[1.1pt]
    \end{tabular}
    \label{tab:urfunny-text-prompt}
\end{table*}

\end{document}